\documentclass[letterpaper, 10 pt, conference]{ieeeconf}  

\IEEEoverridecommandlockouts                              

\usepackage{soul}

\usepackage{graphics} 
\usepackage{svg}
\usepackage{amsmath} 
\usepackage{amssymb}  
\usepackage{comment}
\usepackage{multirow} 
\usepackage{booktabs}

\title{\LARGE \bf
Uncertainty-Aware Shape Estimation of a Surgical Continuum Manipulator in Constrained Environments using Fiber Bragg Grating Sensors
}

\author{Alexander~Schwarz*$^{1,2}$,~\textit{Member,~IEEE}, Arian~Mehrfard*$^{1,2}$,~\textit{Graduate~Student~Member,~IEEE},\\ Golchehr~Amirkhani$^{1}$, Henry~Phalen$^{1}$,~\textit{Graduate~Student~Member,~IEEE}, Justin~H.~Ma$^{1}$,\\~\textit{Graduate~Student~Member,~IEEE}, Robert~B.~Grupp$^{1}$, Alejandro~Martin~Gomez$^{1}$,~\textit{Member,~IEEE}, and \\ Mehran~Armand$^{1}$,~\textit{Member,~IEEE}
\thanks{This work was funded in part by NIH R01EB016703 and NIH R01AR080315 and Johns Hopkins University internal funds.}
\thanks{*Contributed equally to this work and are regarded as joint first authors.}
\thanks{$^{1}$A. Schwarz, A. Mehrfard, G. Amirkhani, H. Phalen, J. Ma, R. Grupp, A. Martin Gomez, and M. Armand are with the Biomechanical- and Image-Guided Surgical Systems (BIGSS) Lab within the LCSR at the Whiting School of Engineering, Johns Hopkins University, Baltimore, MD, USA.
        {\tt\small \{aschwa71, arian.mehrfard, gamirkh1, henry.phalen, jma60, grupp, alejandro.martin, marmand2\}@jhu.edu}}%
\thanks{$^{2}$A. Schwarz and A. Mehrfard are with the Technical University of Munich, Munich, Germany.
        {\tt\small \{alexander.schwarz, arian.mehrfard\}@tum.de}}%
}

\begin{document}

\maketitle
\thispagestyle{empty}
\pagestyle{empty}

\begin{abstract}
Continuum Dexterous Manipulators (CDMs) are well-suited tools for minimally invasive surgery due to their inherent dexterity and reachability. Nonetheless, their flexible structure and non-linear curvature pose significant challenges for shape-based feedback control. The use of Fiber Bragg Grating (FBG) sensors for shape sensing has shown great potential in estimating the CDM's tip position and subsequently reconstructing the shape using optimization algorithms. This optimization, however, is under-constrained and may be ill-posed for complex shapes, falling into local minima. 
In this work, we introduce a novel method capable of directly estimating a CDM's shape from FBG sensor wavelengths using a deep neural network. 
In addition, we propose the integration of uncertainty estimation
to address the critical issue of uncertainty in neural network predictions. Neural network predictions are unreliable when the input sample is outside the training distribution or corrupted by noise. Recognizing such deviations is crucial when integrating neural networks within surgical robotics, as inaccurate estimations can pose serious risks to the patient. We present a robust method that not only improves the precision upon existing techniques for FBG-based shape estimation but also incorporates a mechanism to quantify the models' confidence through uncertainty estimation. We validate the uncertainty estimation through extensive experiments, demonstrating its effectiveness and reliability on out-of-distribution (OOD) data, adding an additional layer of safety and precision to minimally invasive surgical robotics.
\end{abstract}
\section{INTRODUCTION}
\begin{figure}
    \centering
    \includegraphics[width=0.46\textwidth]{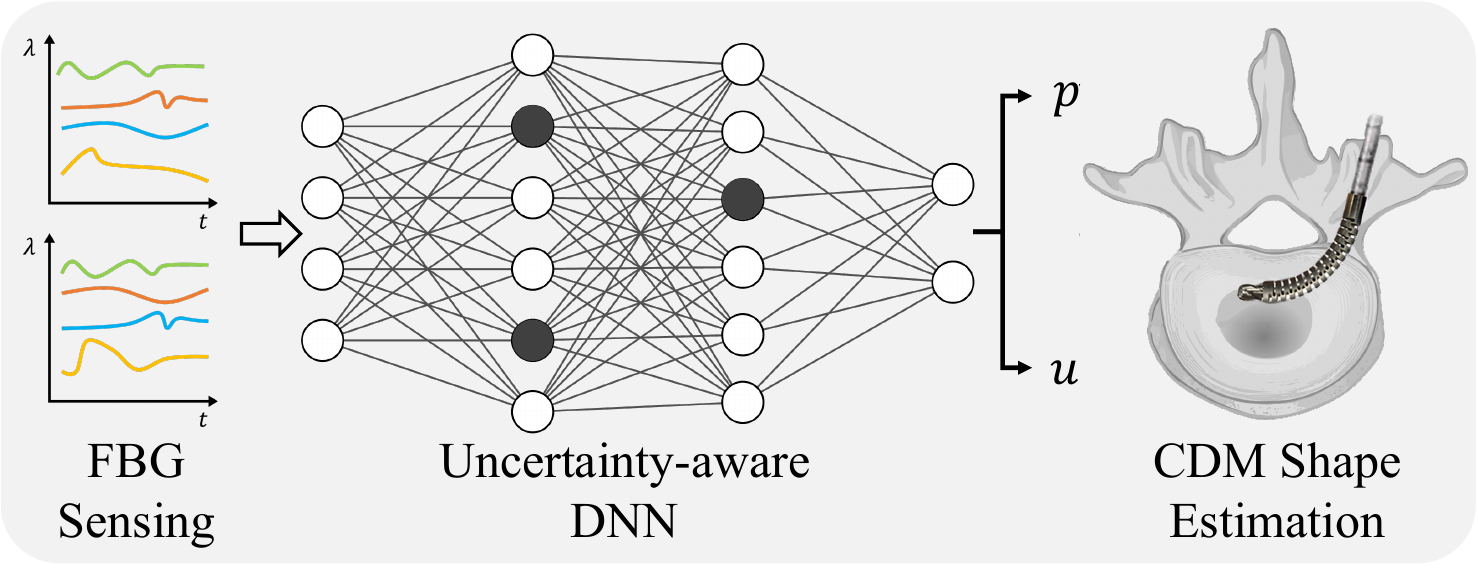}
    \caption{Overview of the proposed method. The wavelength shift $\lambda$ of FBG sensors mounted inside the CDM is measured over time $t$. Mode-corrected wavelength peaks are fed into an uncertainty-aware DNN to  predict the shape of the CDM $\mathbf{p}$ and estimate uncertainty $\mathbf{u}$ for each of the predictions.}
    \label{fig:paper_overview}
\end{figure}
Continuum Dexterous Manipulators (CDMs) have recently demonstrated great potential for minimally invasive orthopedic surgeries due to their flexibility and reach \cite{2017osteonecrosiscdm}. Traditional tools often lack the dexterity for procedures like core decompression for osteonecrosis or precise disk removal in spinal decompression surgery \cite{2017osteonecrosiscdm,wong2014mitlif}. 
To achieve optimal access and navigation through the patient's anatomy, cable-driven CDMs have been proposed (see Fig.~\ref{fig:paper_overview}) \cite{sefati2016fbg}. Those CDMs are designed to accommodate debriding tools and are suitable for both standalone and handheld robotic applications \cite{sefati2021surgrobcdm, ma2021active}.
Due to the inherently non-linear motion of such CDMs, sensing is required to estimate the pose and shape of the robot for a closed control loop. CDM pose can be difficult to model, particularly when interacting with the environment \cite{sefati2016fbg}. As a result, accurate and reliable shape sensing is essential, particularly in the presence of critical anatomical structures during surgery. 

Direct observation of CDM shape is not feasible during many minimally-invasive procedures as there is often little or no line of sight to the tool when inside the human body. In surgical procedures, CDM's shape can be estimated by direct observation via X-ray imaging \cite{vandini2015visionshape,2022congfluorocdm}. However, 
due to the inherent nature of projection images, X-ray imaging can introduce pose ambiguities, complicating the determination of the CDM's shape. This further exposes clinicians and patients to radiation and requires additional equipment, image registration, and post-processing. 
Internal sensing methods offer distinct advantages over radiation-based techniques. Electromagnetic position sensors and optical strain-based sensing approaches, such as Fiber Bragg Grating (FBG) sensors, have been explored \cite{sadjadi2016simultaneous, hill1997fiber, ryu2014fbg, monet_2020_icra_survey, amirkhani2023design}. For orthopedic procedures, size constraints and the introduction of electromagnetic noise in the presence of ferromagnetic instruments \cite{sadjadi2016simultaneous,franz2014emreview, feuerstein2009magneto}, such as a rotating burr during drilling tasks, make FBG sensors favorable \cite{sefati2020data}. 

FBG sensors provide a radiation-free, high-frequency stream of strain data along the length of the CDM, which can be used to estimate the shape of the CDM.
The conventional use of FBG fibers with an interrogator allows the strain at the FBG nodes to be used for shape estimation of the fiber.
One approach computes a single fiber's shape using curvature, torsion, and Frenet-Serret equations \cite{moore2012shape}.
This approach was extended to facilitate the shape reconstruction of a medical catheter using multiple fibers \cite{khan2019multi}. Another established FBG shape sensing method assumes each segment to have a constant curvature and reconstructs the CDM shape by concatenating discrete arcs of the estimated curvature at each sensor node location \cite{roesthuis2016steering}. 
Deaton \textit{et al.}, \cite{deaton2023towards}, employed a constant curvature model, integrating data across three fiber sections to reconstruct the manipulator's shape. They further introduced a simultaneous 3D shape and force sensing assembly, which was incorporated into a planar continuum manipulator designed for medical applications \cite{deaton2023simultaneous}.

Another method trains machine learning models on the FBG sensor readings to obtain the shape of the CDM which is often referred to as data-driven shape sensing. Prior work on data-driven shape sensing fitted different models to predict the distal end position of the CDM and reconstruct the shape through a constrained optimization problem \cite{sefati2020data}. Given the high flexibility of continuous systems, however, this optimization is under-constrained and may be ill-posed for complex shapes, falling into local minima. 

In this work, we directly estimate the full CDM shape from FBG sensor readings with a deep neural network (DNN) eliminating the under-constrained optimization for complex shapes. We utilize a sensor assembly of two optical fibers with four FBG nodes each. The neural network is supervised with a shape ground truth that is extracted from thirty markers placed along the CDM centerline. The shape estimation is evaluated on large deflections of the CDM in free and constrained space through obstacles. 

Yet, an essential concern for data-driven techniques to shape estimation in medical applications remains. Inaccurate predictions of the CDM shape can potentially lead to catastrophic damage to the patient's anatomy. Domain shift between the experimental setup used for training and the actual conditions during surgery poses an essential challenge for data-driven shape estimation and machine learning in robotics. \cite{loquercio_general_2020}. Addressing this, we propose an uncertainty-aware approach to direct shape estimation, that attempts to not only predict the shape of the CDM via a DNN but also estimate the neural network's confidence in its prediction via Monte Carlo Dropout.
Our primary contributions are:
\begin{itemize}
    \item A method that employs a neural network to directly estimate the shape of a surgical CDM using a novel FBG sensor array consisting of two fibers with four nodes each.
    \item Uncertainty estimation for sensor-based shape reconstruction, which offers a mechanism to gauge the neural network's confidence in its predictions.
    \item Comprehensive testing and validation of both shape and uncertainty estimation through rigorous experiments of dynamic left and right side bending in freespace and in constrained environments.
\end{itemize}

\section{METHODS}
\subsection{FBG-based Shape Sensing}
Relative to a chosen reference state, often the neutral position with no strain, the shift in wavelength for each FBG node of a fiber is linearly dependent on the applied mechanical strain $(\epsilon_x)$ and the change in temperature $(\Delta T)$ \cite{roesthuis2013three}: 
\begin{equation} \label{eq:1} 
    \Delta\lambda_B = \lambda_B ( (1-p_e)\epsilon_x + (\alpha_{\Lambda} + \alpha_n)\Delta T)
\end{equation}
where $\Delta\lambda_B$ is the shift in FBG wavelength, $\lambda_B$ is the wavelength of light reflected by the FBG in its reference state, and $p_e$ is the photoelastic coefficient of the optical fiber. The thermal expansion coefficient $\alpha_{\Lambda}$ and the thermo-optic coefficient $\alpha_n$ describe the changes in the length of the optical fiber and the changes in the refractive index of the fiber with respect to changes in temperature.
The effect of temperature on the wavelength shift can be compensated for by subtracting the common mode from the wavelength shift ($\Delta\lambda_B$) of two or more axisymmetric fibers \cite{iordachita2009sub,gonenc2017towards}. Assuming a negligible temperature gradient, the common mode is computed as the mean wavelength shift of two adjacent active areas. Without the influence of temperature, Eq.~\ref{eq:1} can be rewritten to illustrate the linear correlation between FBG wavelength shift and mechanical strain:
\begin{equation} \label{eq:2} 
    \epsilon_x = \frac{\Delta \lambda_B}{\lambda_B(1-p_e)}
\end{equation}

Conventional, model-based shape reconstruction methods use calibrated sensors with known geometrical models to estimate curvature and bending direction at the active areas of the optical fiber and then integrate over its length to estimate the tip position \cite{sefati2019fbg}. Recent works use a data-driven approach to estimate the tip position of the CDM from FBG wavelengths and a constrained optimization to reconstruct the shape of the CDM from the tip position \cite{sefati2019fbg, sefati2020data}.

\subsection{Data-Driven Shape Estimation}
\begin{figure}
    \centering
    \includegraphics[width=0.4\textwidth]{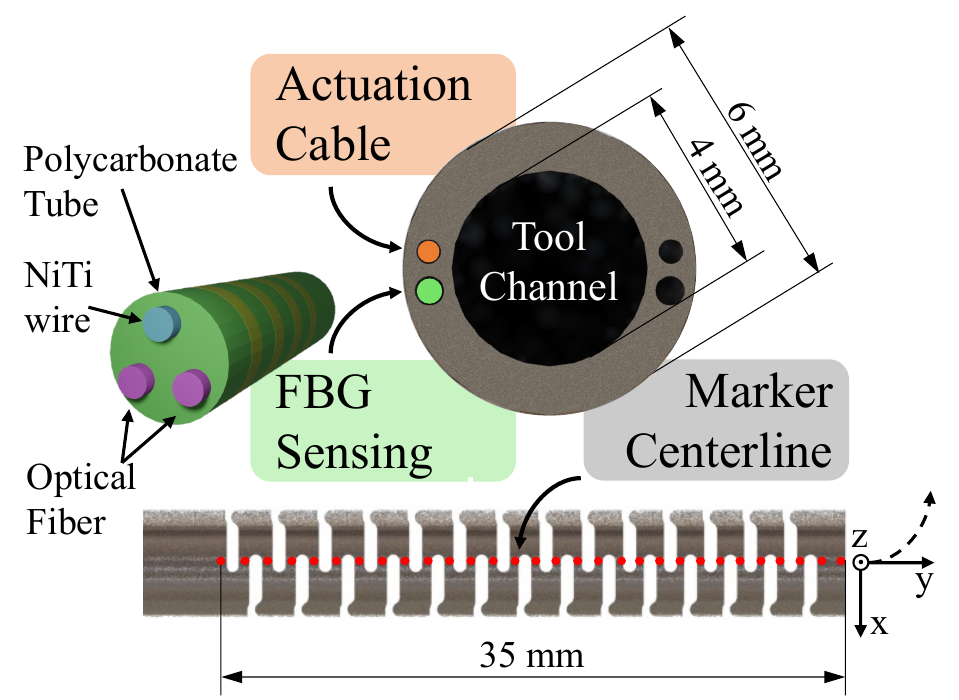}
    \caption{Schematic cross-sectional and top-down view of the CDM showing the actuation cable, FBG sensing, and tool channel. The FBG sensing array consists of two optical fibers and a NiTi wire embedded in a polycarbonate tube and is inserted into the CDMs wall (green) below the actuation cable (orange). The CDM bends within the $(x,y)$-plane. Red markers are placed along the centerline for ground truth shape reconstruction.}
    \label{fig:cdm_schematic}
\end{figure}

This work proposes a purely data-driven method for the shape estimation of CDMs eliminating prior geometric or structural assumptions, calibrations, and models. We utilize a DNN to directly estimate the shape of a CDM from FBG sensing data. For this purpose, our sensing unit is equipped with two optical fibers. Each fiber has four active areas spaced 8 mm apart. Additionally, a nitinol wire is embedded within the polycarbonate tube (see Fig.~\ref{fig:cdm_schematic}) \cite{amirkhani2023design}. Its memory shape properties can be disregarded as any change in the CDM's shape will be reflected in the FBG measurements. The CDM we use is a Nitinol tube with notches along its length that allow planar bending while remaining stiff in the direction perpendicular to the bending plane. The CDM has an outer diameter of $6$~mm and an inner diameter of $4$~mm that serves as an instrument channel. Two pairs of channels are present in the bendable walls of the CDM. Each pair consists of one $0.5$~mm diameter channel for the actuation cable and a sensing channel with a diameter of $0.6$~mm, which contains the FBG sensing unit. The CDM design is identical to the one described in \cite{ma2021active}, with the only difference that thirty color markers were added along its centerline, as shown in Fig.~\ref{fig:cdm_schematic}. Each marker was allocated to a unique segment, ensuring complete coverage of the CDM's dexterous regions. A stereo camera configuration captures the color markers to establish a precise ground truth representation of the CDM's shape (see~Fig.~\ref{fig:experiment_setup}). The input to our data-driven model consists of eight mode-corrected FBG wavelength peaks from the two optical fibers. We supervise against the corresponding ground truth shape represented by thirty discrete markers, each expressed in millimeters as \( (p^{i}_{x}, p^{i}_{y}) \) for \( i = 1,...,30 \). These denote the \( x \) and \( y \) coordinates of the markers in the left camera coordinate system with the origin set at the first marker's location \( (p^{1}_{x}, p^{1}_{y}) \).
\subsection{Neural Network Architecture}
We train the neural network $f$ as, $\mathbf{Y} = f_{\boldsymbol{\theta}}(\mathbf{X})$, with neural network weights $\boldsymbol{\theta}$, to predict the distinct marker locations $\mathbf{Y}$ given the processed sensor readings $\mathbf{X}$.
The proposed network architecture is a fully connected Multi-Layer Perceptron (MLP), consisting of three layers with $512$, $256$, and $60$ neurons in each layer, respectively. We use Rectified Linear Unit (ReLU) activation functions and the Adaptive Moment Estimation (ADAM) optimizer for training. 
The architecture distinguishes itself from previous work in \cite{sefati2020data} mainly through the incorporation of Dropout, with dropout probability $d$, applied to each layer of the MLP. This introduces regularization to mitigate overfitting and enables to estimate the uncertainty associated with its predictions. Mean Squared Error (MSE) loss is used for both Tip Point Estimation (TPE) and Direct Shape Estimation (DSE).
The input data is normalized using min-max-normalization for each feature individually within the training data. The normalization parameters are retained and subsequently applied to the test set in an analogous manner.

\subsection{Uncertainty Estimation}
The safe application of neural networks to shape estimation in surgical robotics necessitates the system to not only provide a prediction but also offer a confidence metric about the model's prediction. This confidence measure can then be utilized by the surgeon or higher level system to validate the reconstruction through the means of other, more costly, shape estimation modalities, like X-ray imaging \cite{2022congfluorocdm}. We address this by estimating the uncertainty of our model, often referred to as model or epistemic uncertainty. This uncertainty can be characterized by placing a distribution over the neural network's weights, $\boldsymbol{\theta}$. The weight distribution can be expressed as $p(\boldsymbol{\theta}|\mathbf{X},\mathbf{Y})$, where $\mathbf{X}$ are the training samples and $\mathbf{Y}$ the target values. $p(\boldsymbol{\theta}|\mathbf{X},\mathbf{Y})$ is intractable for most cases. To approximate this distribution, we employ Monte Carlo Dropout by collecting weight samples while using dropout during inference \cite{gal_dropout_2016}. This approximates:
\begin{equation}
   p(\boldsymbol{\theta}|\mathbf{X},\mathbf{Y}) \approx q(\boldsymbol{\theta}; \mathbf{d}) = \textit{Bern}(\boldsymbol{\theta}; \mathbf{d})
\end{equation}
where $\mathbf{d}$ are the dropout rates on the neural network weights. Given this assumption, the model uncertainty is estimated as the variance of $K$ Monte-Carlo samples \cite{loquercio_general_2020}:
\begin{equation}
    \mathbf{\sigma} = \frac{1}{K} \sum_{k=1}^{K} (\mathbf{y}_k - \overline{\mathbf{y}})^2
\end{equation}
where $\{\mathbf{y}_k\}_{k=1}^{K}$ is a set of $K$ sampled predictions for weights instances $\boldsymbol{\theta}^k \sim q(\boldsymbol{\theta}; \mathbf{d})$ and $\overline{\mathbf{y}} = \frac{1}{K}\sum_{k}{\mathbf{y}_k}$. Monte Carlo Dropout replicates an ensemble of different models at inference time and can be explained intuitively. When multiple different models all predict the same shape, the uncertainty is low thus the confidence is high. When the different models predict different shape values for the same input, the confidence is low. 
We leverage model uncertainty estimates $\boldsymbol{\sigma}$ to define a confidence interval $\mathbf{u} = \omega * \boldsymbol{\sigma}$, with scaling factor $\omega$, for each marker prediction separately. 

\section{EXPERIMENTS}
\subsection{Experimental Setup}
\begin{figure}[b]
    \centering
    \includegraphics[width=0.47\textwidth]{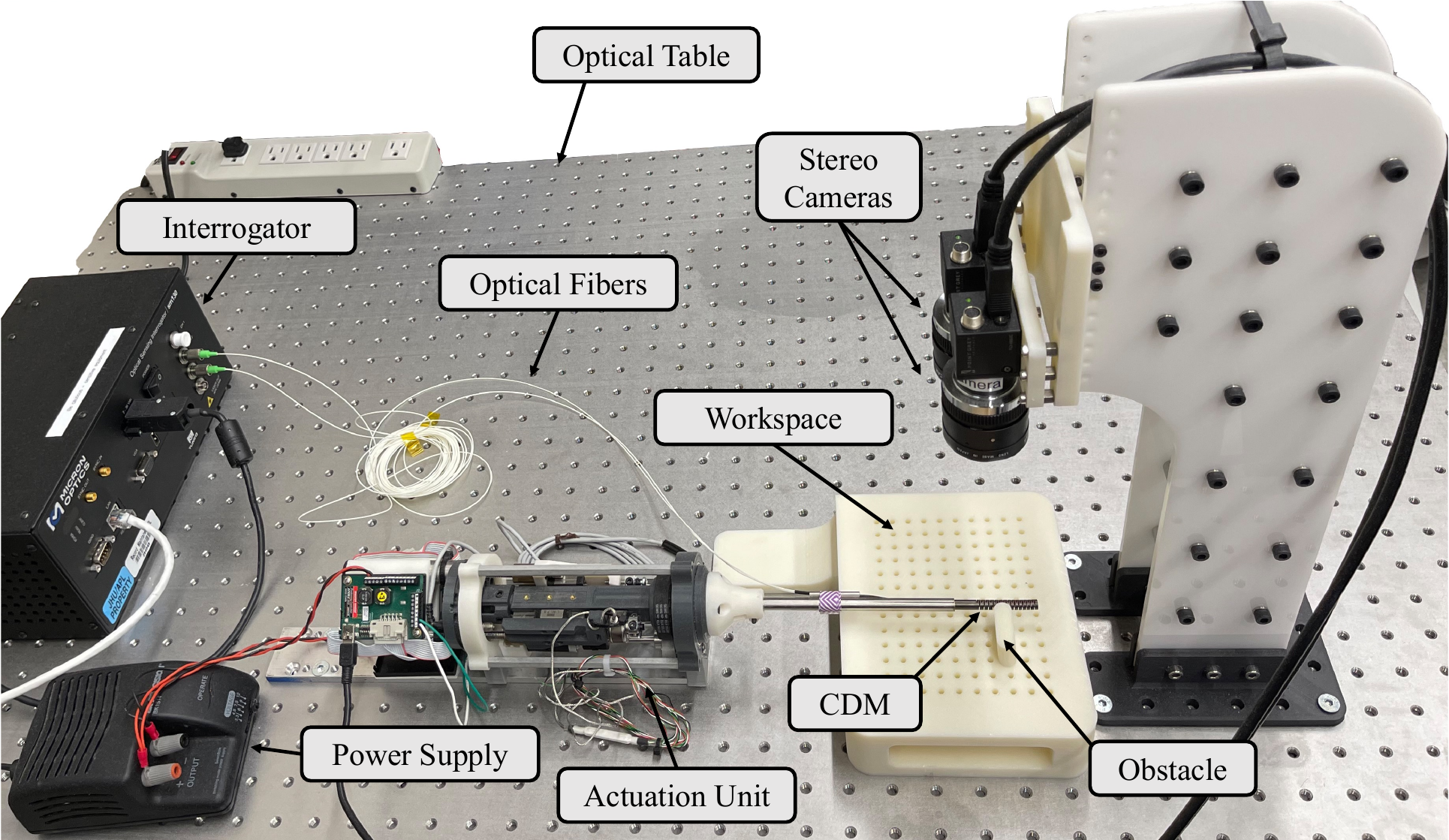}
    \caption{Experimental Setup including, the stereo cameras, CDM, actuation unit and FBG sensing unit mounted on the optical table.}
    \label{fig:experiment_setup}
\end{figure}

We used a test-bench setup as shown in Fig. \ref{fig:experiment_setup} \cite{amirkhani2023design}. The CDM was attached to an actuation unit with linear motors (4.5W, RE 16, Maxon, Switzerland) to pull cables embedded in the left or right wall of the CDM and fixated at the CDM tip, (see Fig.~\ref{fig:cdm_schematic}). The cable displacement leads to planar bending in the direction corresponding to the location of the cable. FBG data was collected via an optical sensing interrogator (Micron Optics sm 130, Luna Inc., Atlanta, GA, USA) and streamed to the computer at $100$~Hz . Two cameras (Flea2 1394b, FLIR Integrated Imaging Solutions Inc., Wilsonville, OR, USA) were placed in stereo arrangement $30$~cm above the CDM working area streaming images at $15$~Hz to collect ground truth data. 
To generate diverse training and test sets, we collected extensive data on the real system, both in freespace and in constrained environments. In the freespace experiments, the CDM can bend without obstruction.

\begin{figure}[t]
    \centering
    \includegraphics[width=0.45\textwidth]{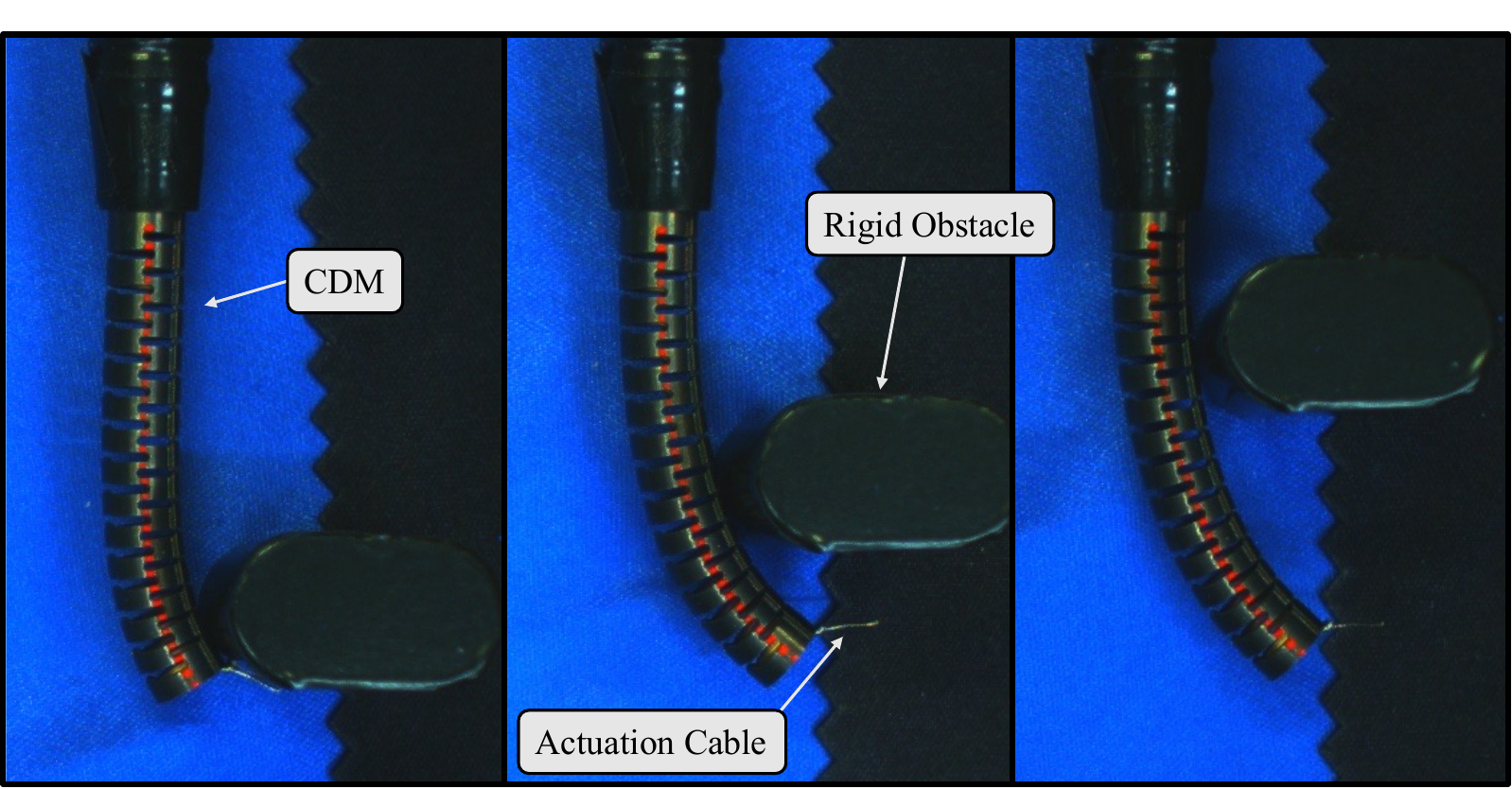}
    \caption{CDM bending experiments with obstacles. Rigid obstacles were placed at the tip position (left), center (middle), and base (right) of the CDM.}
    \label{fig:obstacles_experiment_setup}
\end{figure}

To simulate contact forces along the CDM we ran experiments in constrained environments. For this purpose, obstacles with an outer radius of $10$~mm were 3D printed and rigidly placed at six different locations. The obstacles deflect the CDM from its ability to bend freely resulting in complex shapes. The obstacles were placed near the distal tip, the center segment, or the proximal end of the CDM, both on the left and the right side of the CDM as shown in Fig.~\ref{fig:obstacles_experiment_setup}. For each of the obstacle placements, data was recorded separately. In both constrained and unconstrained experiments, the CDM was actuated from a straight neutral position with $0$~mm cable displacement to a strong deflection with a maximum of $5$~mm cable displacement, which equates to $81^{\circ}$ angular deflection of the tooltip compared to the neutral position in freespace. To simulate cable velocities at which the cutting tool is fully engaged in the tissue \cite{ma2021active}, the cable was pulled at velocities ranging from $0.1$ to $0.4$~mm/s.  

\subsection{Data Collection}
To achieve concurrent and synchronized data collection from the FBG sensors and cameras, we developed a thread-safe application using three CPU threads to collect, timestamp, and store the FBG wavelengths and camera frames at their respective frequencies. In post-processing, the timestamps were used to pair FBG and camera samples, achieving an average time difference of $3.6$~ms between camera and FBG data.
We collected a total of 61,999 data points for training, equating to 59 bends, and conducted separate experiments, yielding 38,400 data points for testing. To assess the uncertainty estimation, our model was not trained on two out of the six obstacle placements: left-side bending with obstacles situated at the base and center positions. This introduces a distribution shift, as the model was exposed to scenarios during testing that it had not encountered during training. We refer to this data as out-of-distribution (OOD) data, representing 6,678 samples out of our test set.

\subsection{Ground Truth Reconstruction}
\begin{figure}[h]
    \centering
    \includegraphics[width=0.4\textwidth]{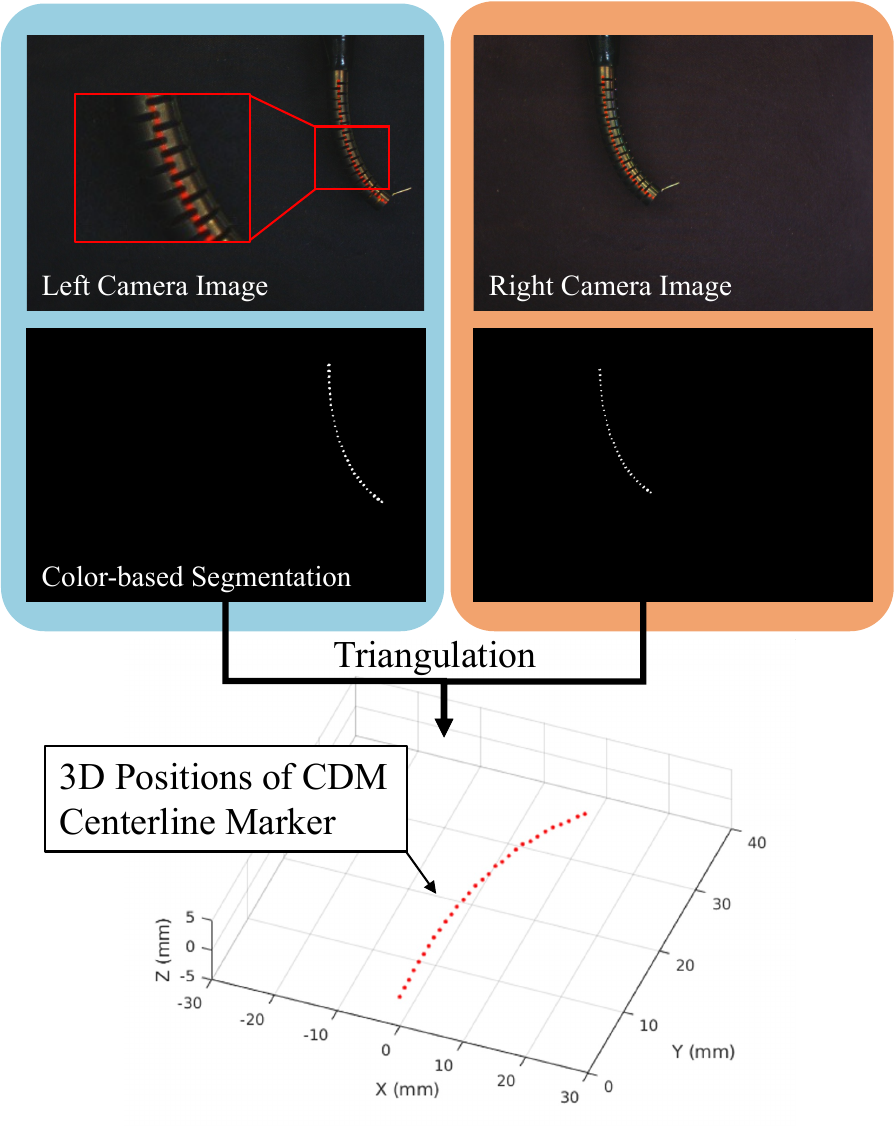}
    \caption{3D Shape Ground Truth Extraction of the CDM centerline from stereo images using color-based segmentation and triangulation of the detected marker centroids.}
    \label{fig:stereomatching_schema}
\end{figure}

Stereo cameras were rigidly mounted over the test bench and used to track the red markers attached to the centerline of the CDM, as shown in Fig. \ref{fig:experiment_setup}. 
We calibrated the $1024\times768$ stereo camera pair using the MATLAB stereo camera calibration toolbox, obtaining both intrinsic and extrinsic matrices. The stereo camera calibration resulted in an average reprojection error of $0.23$~pixels. The algorithm for recovering the shape of the CDM is visualized in Fig.~\ref{fig:stereomatching_schema}. Color-based segmentation was applied to the left and right images to obtain a segmentation mask for subsequent centroid detection. We triangulated the positions of the thirty detected centroids, resulting in the 3D positions of the CDM centerline that we used as ground truth for training and testing our models. Any sample where more than one of the thirty markers was identified as an outlier was excluded from the dataset, accounting for less than 4\% out of all samples. Conversely, when only a single marker was flagged as an outlier, it was substituted with a synthetic marker, inserted under the assumption of constant curvature.
\section{RESULTS}
A summary of the shape estimation performance of our data-driven method is shown in Tab.~\ref{tab:results_tpe_error} for tip point estimation and in Tab.~\ref{tab:results_dse_error} for direct shape estimation. We compare our method against two baselines, which are a linear regression model and a polynomial regression model of second order. The DNN was trained with dropout rate $d=0.3$ and for evaluation $K=100$ inference calls were made with dropout enabled. Given this setup, we can run the model at $11.1$~Hz on a consumer CPU (12th Gen Intel Core i7-12700K). Errors are reported in euclidean distance between the predicted marker position and the ground truth target. Metrics were computed over multiple continuous, dynamic bends from the neutral position to a maximum of $5$~mm cable displacement. We report both, the errors for freespace bending, as well as constraint environments, with obstacles placed at different locations. The maximum shape estimation error is defined by the highest individual marker error.
Since the tip point consistently exhibits the highest error, maximum errors for shape and tip estimation align.
The models were not trained on the \textit{Center Left} and \textit{Tip Left} obstacle placements, as those setups are used to evaluate OOD performance. 
\begin{table}[b]
  \centering
  \caption{CDM Tip Point Estimation Errors for bending in freespace and in constrained environments.}
  \begin{tabular}{l|ccc|ccc} 
    \toprule
    \multicolumn{1}{c}{} & \multicolumn{3}{c}{Median Error [mm]} & \multicolumn{3}{c}{Max Error [mm]} \\
    \cmidrule(lr){2-4} \cmidrule(lr){5-7} 
    & Lin & Poly & DNN & Lin & Poly & DNN \\
    \midrule
    \textbf{Freespace}     & 0.251 & 0.124 & \textbf{0.085} & 1.10 & 1.028 & \textbf{1.021} \\
    \midrule
    \textbf{Obstacles}     & 0.283 & 0.116 & \textbf{0.098} & 3.043 & 3.365 & \textbf{2.180} \\
    Base Right    & 0.173 & 0.108 & \textbf{0.063} & 0.972 & 0.488 & \textbf{0.462} \\
    Center Right  & 0.217 & 0.131 & \textbf{0.083} & \textbf{0.937} & 3.365 & 1.574 \\
    Tip Right     & 0.215 & 0.089 & \textbf{0.059} & 1.105 & 0.950 & \textbf{0.803} \\
    Base Left     & 0.485 & \textbf{0.142} & 0.230 & 1.494 & \textbf{0.658} & 0.673 \\
    Center Left   & 0.439 & \textbf{0.106} & 0.198 & 3.043 & \textbf{1.324} & 2.169 \\
    Tip Left      & 0.379 & 0.212 & \textbf{0.195} & 1.548 & \textbf{0.611} & 0.722 \\
    \bottomrule
  \end{tabular}
  \label{tab:results_tpe_error}
\end{table}
We find that the DNN achieves the smallest median error as well as maximum error in most experiments. Linear regression achieves the worst accuracy in most of the experiments. We further find that the error is the largest with the obstacle placed at the center position. In freespace, the DNN achieves a median shape estimation error of $0.031$~mm while the maximum error does not exceed $1.021$~mm. In constrained environments, the DNN still performs best for most cases, with a median error of $0.058$~mm for shape estimation and a maximum error of $2.180$~mm.
\begin{table}[b]
    \caption{CDM Shape Estimation Errors for bending in freespace and in constrained environments.}
    \centering
    \begin{tabular}{l|ccc|ccc} 
    \toprule
    \multicolumn{1}{c}{} & \multicolumn{3}{c}{Median Error [mm]} & \multicolumn{3}{c}{Max Error [mm]} \\
    \cmidrule(lr){2-4} \cmidrule(lr){5-7} Experiment
    & Lin & Poly & DNN & Lin & Poly & DNN \\
    \midrule
    \textbf{Freespace}     & 0.063 & 0.041 & \textbf{0.031} & 1.10 & 1.028 & \textbf{1.021} \\
    \midrule
    \textbf{Obstacles}     & 0.083 & 0.069 & \textbf{0.058} & 3.043 & 3.365 & \textbf{2.180}\\
    Base Right    & 0.085 & 0.074 & \textbf{0.058} & 0.972 & 0.488 & \textbf{0.462} \\
    Center Right  & 0.078 & 0.077 & \textbf{0.051} & \textbf{0.937} & 3.365 & 1.574\\
    Tip Right     & 0.078 & 0.062 & \textbf{0.053} & 1.105 & 0.950 & \textbf{0.803} \\
    Base Left     & 0.074 & 0.066 & \textbf{0.051} & 1.494 & \textbf{0.658} & 0.673 \\
    Center Left   & 0.088 & \textbf{0.058} & 0.071 & 3.043 & \textbf{1.324} & 2.169 \\
    Tip Left      & 0.159 & \textbf{0.108} & 0.123 & 1.548 & \textbf{0.611} & 0.722 \\
    \bottomrule
    \end{tabular}
    \label{tab:results_dse_error}
\end{table}
\begin{figure}[ht]
    \centering
    \includegraphics[width=0.48\textwidth]{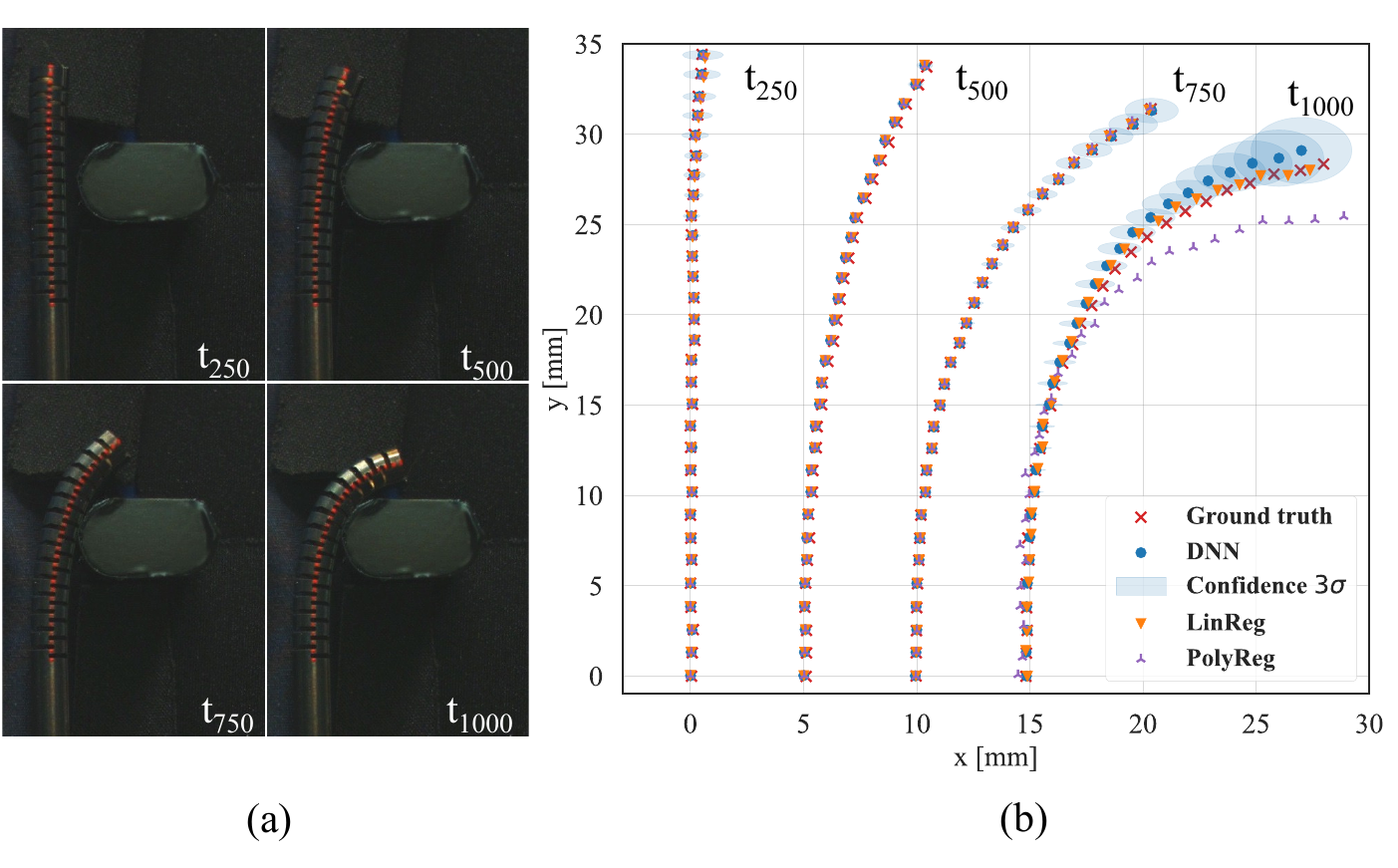}
    \caption{Qualitative shape estimation results for the DNN with confidence intervals of $3 \sigma$, Linear Regression, and Polynomial Regression at four bending stages with an obstacle placed at the center. (a) shows the left camera image. (b) displays shape estimation results and confidence intervals. Ground truth was extracted from the stereo images. Plots are shifted by $5$~mm in $x$ direction to avoid superposition.}
    \label{fig:qualitative_shape_reconstruction}
\end{figure}
\begin{figure*}[tb]
    \centering 
    \includegraphics[width=0.99\textwidth]{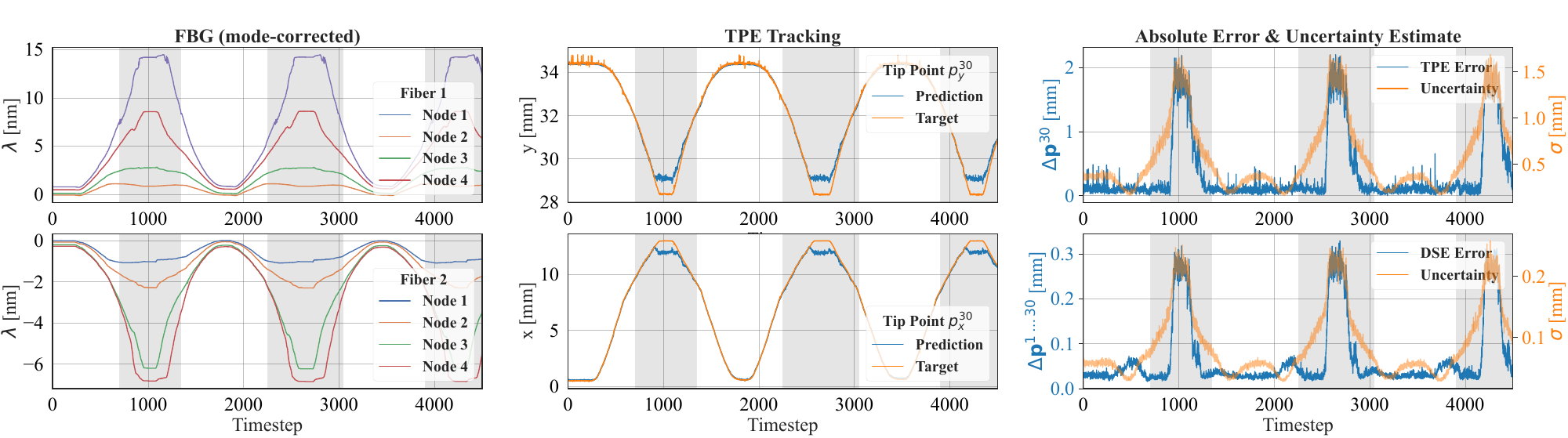}
    \caption{Results of experiments involving multiple right-side bends of the CDM with a central obstacle. The left panel displays FBG measurements for two fibers at four nodes post mode-correction. The different peak amplitudes are a result of the non-uniform bending along the fiber. Tension increases the wavelength while compression reduces it, leading to positive peaks for Fiber 1 and negative peaks for Fiber 2 post mode-correction. 
    The middle panel shows TPE tracking in x and y positions. The right panel indicates error, complemented by uncertainty estimates for both TPE and DSE over time. Grey-shaded intervals denote when the CDM contacts the obstacle. Notably, larger uncertainty estimates correspond with higher error in bending shapes.}
    \label{fig:tracking_results}
\end{figure*}
Given the largest errors occur at center-located obstacles, Fig.~\ref{fig:tracking_results} shows the tracking of multiple right-side bends over time for this setup. Notably, the prediction $\overline{p}$ deviates from the ground truth ${p}$ around the 1000th timestep when bending is particularly strong. This results in a higher error $\Delta{p}$. Our method detects this deviation with increased uncertainty estimates. The uncertainty estimates show a smoother and more consistent trend compared to the error values. Fig.~\ref{fig:qualitative_shape_reconstruction} displays the shape estimation results at four distinct timesteps of the same experiment, ranging from no cable displacement to an angular deflection of $71^{\circ}$. Fig.~\ref{fig:qualitative_shape_reconstruction} (a) shows the corresponding images of the left camera, used to extract ground truth. The estimated shapes from the DNN, Linear Regression, and Polynomial Regression are visualized together with the ground truth in Fig.~\ref{fig:qualitative_shape_reconstruction} (b). The confidence interval $3\sigma$ for the neural network's prediction stemming from the uncertainty estimates for each $x$ and $y$ marker position predictions are visualized as ellipses. While Polynomial Regression falls short in accurately estimating the shape, Linear Regression achieves the lowest error. Meanwhile, any deviation of the DNN from the ground truth is consistently reflected in heightened uncertainty estimates.

\begin{figure}[h]
    \centering
    \includegraphics[width=0.5\textwidth]{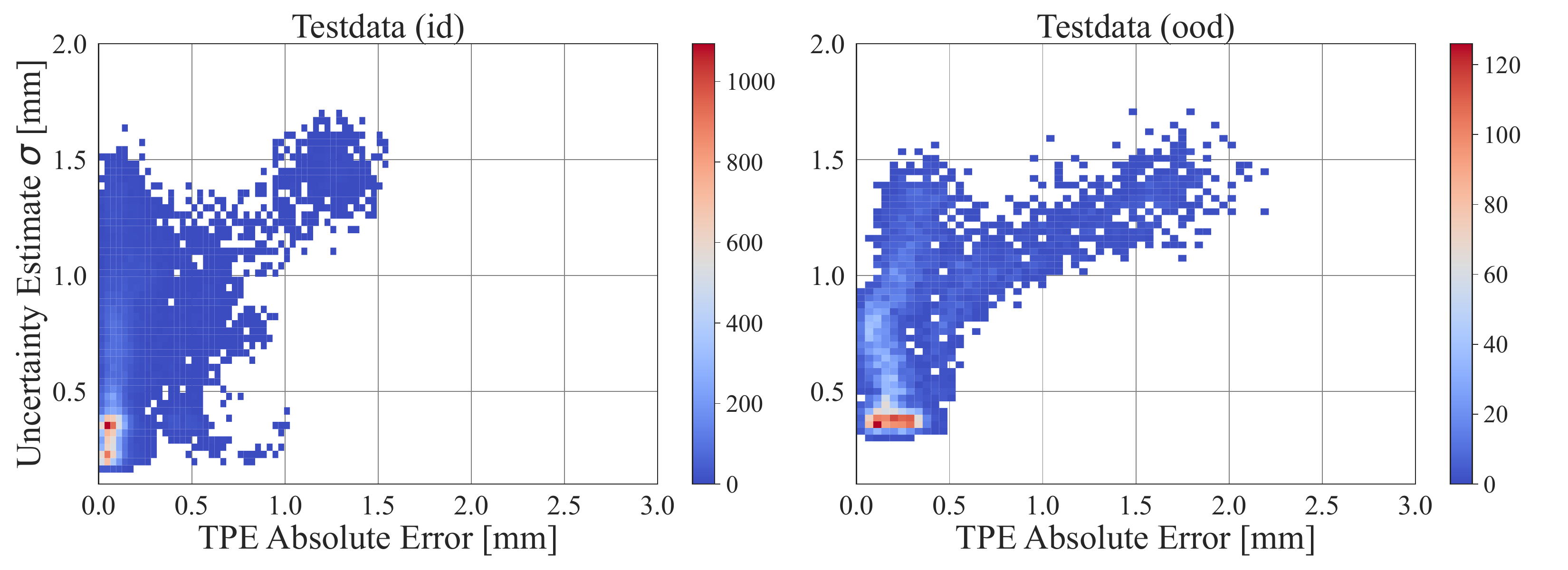}
    \caption{Distribution of the TPE Error in relation to Uncertainty Estimates $\sigma$ for the test datasets, in-distribution (id) and out-of-distribution (ood). Samples in the bottom right would indicate False Positives, representing a critical area where the system exhibits high confidence in predictions with significant errors. We do not report any Absolute Errors larger than $1.5$~mm for uncertainty estimates with $\sigma<1.0$~mm.}
    \label{fig:uncertainty_distribution}
\end{figure}

We further analyze the effectiveness of the proposed method for uncertainty estimation on our two testing distributions, in-distribution and out-of-distribution. Ideally, the error linearly correlates with the uncertainty estimate $u$. False positive values, however, are the most critical as the system would exhibit high confidence in predictions with significant errors. We present the distribution of the TPE error in relation to the uncertainty estimates in Fig.~\ref{fig:uncertainty_distribution}. 
Separate plots display the testing data stemming from the same distribution and the one we categorize as OOD. Notably, we observe no Absolute Errors exceeding $1.1$~mm for uncertainty estimates under $\sigma<1.0$~mm. For both data distributions, most of the samples are concentrated in a region where the uncertainty is $\sigma<0.5$~mm and the Absolute Error is less than $0.5$~mm, as evidenced by the predominant red areas. 
\section{DISCUSSION}
Our experimental outcomes highlight the potential of using an uncertainty-aware deep neural network for shape sensing of CDMs from FBG sensor readings. Specifically, our method, when compared to traditional approaches such as linear and polynomial regression, achieved lower errors, both in terms of median and maximum error metrics. Importantly, our method represents the first instance where a data-driven model addresses both left and right side bending in dynamic environments, thereby further proving the applicability of FBG-based shape sensing of CDMs for surgical environments. 
We report lower median errors to previous model-based and data-driven approaches for shape sensing with comparable maximum errors \cite{amirkhani2023design,sefati2020data}. While achieving low errors for shape estimation, the deviation from the ground truth increases for the tip point when obstacles are placed at the center of the CDM. In future work, a more balanced training dataset could be designed to ensure even representation across various angular deflections of the tooltip. We introduced a robust method for estimating uncertainty in neural network predictions, which we believe is crucial to the safe application for shape sensing in surgical robotics. We tested our method on unseen environments. Our approach to uncertainty estimation, while proving to be effective, leans towards being conservative. This conservative stance, however, ensures a layer of safety that's paramount in the context of minimally invasive surgical robotics. The estimated confidence level can then be used by the surgeon or higher level system, to decide, when it is beneficial to validate the estimation through the means of other shape estimation modalities, like X-ray imaging. All experiments were run with $K=100$ inference calls. The best choice of $K$ is a trade-off between inference speed and model performance. We found similar performance for $K=25$, which speeds up inference to $44.7$~Hz.
\section{CONCLUSION}
In this study, we introduced a method for directly sensing the shape of a CDM using eight FBG sensor readings through an uncertainty-aware neural network, incorporating Monte Carlo Dropout to address model uncertainty. Our results reliably detect incorrect predictions in both in-distribution and out-of-distribution contexts. Through extensive experiments, we demonstrated our method's capability to accurately estimate the CDM's shape both in freespace and in constrained environments and to identify mispredictions via uncertainty estimation. Uncertainty estimation provides an additional layer of safety and precision to shape reconstruction, making it a viable asset for applying neural networks to minimally invasive surgical robotics. In future work, our goal is to integrate the CDM into the robotic platform and assess shape-sensing performance in phantom and cadaveric experiments. We are further interested in combining FBG-based shape sensing with other modalities such as X-ray imaging through sensor fusion. Additionally, we intend to explore diverse visualization techniques for uncertainty-aware shape estimation of CDMs to discern the information most beneficial to surgeons.

\addtolength{\textheight}{-6cm}   

\bibliographystyle{unsrt}
\bibliography{literature}
\end{document}